\def\year{2019}\relax
\documentclass[letterpaper]{article} 
\usepackage{aaai19}  
\usepackage{times}  
\usepackage{helvet}  
\usepackage{courier}  
\usepackage{url}  
\usepackage{graphicx}  
\usepackage{tabularx}
\usepackage{multirow}
\usepackage{amssymb}
\usepackage{amsmath}
\usepackage{pifont}
\usepackage{booktabs}
\newcommand{\cmark}{\ding{51}}%
\newcommand{\xmark}{\ding{55}}%

\title{DDFlow: Learning Optical Flow with Unlabeled Data Distillation}
\author{
Pengpeng Liu$^\dag$\thanks{Work mainly done during an internship at Tencent AI Lab.}, 
Irwin King$^\dag$, 
Michael R. Lyu$^\dag$, 
Jia Xu$^\S$ \\
$^\dag$ The Chinese University of Hong Kong, Shatin, N.T., Hong Kong \\
$^\S$ Tencent AI Lab, Shenzhen, China \\
\texttt{\{ppliu, king, lyu\}@cse.cuhk.edu.hk, jiajxu@tencent.com }
}

\begin{document}
\maketitle

\begin{abstract}
We present DDFlow, a data distillation approach to learning optical flow estimation from unlabeled data. The approach distills reliable predictions from a teacher network, and uses these predictions as annotations to guide a student network to learn optical flow. Unlike existing work relying on hand-crafted energy terms to handle occlusion, our approach is data-driven, and learns optical flow for occluded pixels. This enables us to train our model with a much simpler loss function, and achieve a  much higher  accuracy. We conduct a rigorous evaluation on the challenging Flying Chairs, MPI Sintel, KITTI 2012 and  2015 benchmarks, and show that our approach significantly outperforms all existing unsupervised learning methods, while running at real time.
\end{abstract}

\section{Introduction}

Optical flow estimation is a core computer vision building block, with a wide range of applications, including  autonomous driving \cite{menze2015object}, object tracking \cite{chauhan2013moving}, action recognition \cite{simonyan2014two} and video processing \cite{Bonneelsiggraph2015}.  Traditional approaches \cite{horn1981determining,brox2004high,brox2011large} formulate optical flow estimation as an energy minimization problem, but they are often computationally expensive \cite{XRK2017}. Recent learning-based  methods \cite{dosovitskiy2015flownet,ranjan2017optical,ilg2017flownet,hui18liteflownet,sun2018pwc} overcome this issue  by  directly estimating optical flow from raw images  using convolutional neural networks (CNNs).
However, in order to train such CNNs with high performance, it requires a large collection of densely labeled data, which is extremely difficult to obtain for real-world sequences.

 One alternative is to use synthetic datasets.
Unfortunately, there usually exists a large domain gap between the distribution of synthetic images and natural scenes \cite{liu2008human}. Previous networks \cite{dosovitskiy2015flownet,ranjan2017optical} trained only on synthetic data turn to overfit, and  often perform poorly when they are directly evaluated on real sequences.
 Another promising direction is to learn from unlabeled videos, which are readily available at much larger scale. \cite{jason2016back,ren2017unsupervised} employ the classical warping idea, and train CNNs with a photometric loss defined on the difference between reference and warped target images. Recent methods propose  additional loss terms  to cope with occluded pixels \cite{Meister:2018:UUL,wang2018occlusion}, or utilize multi-frames to reason occlusion \cite{Janai2018ECCV}.
  However, all these methods rely on hand-crafted energy terms to regularize optical flow estimation,  lacking key  capability to learn optical flow of occluded pixels.  As a result, there is still a large performance  gap comparing these methods with state-of-the-art fully supervised methods.

 \begin{figure}
 \centering
 \includegraphics[width=0.45\textwidth]{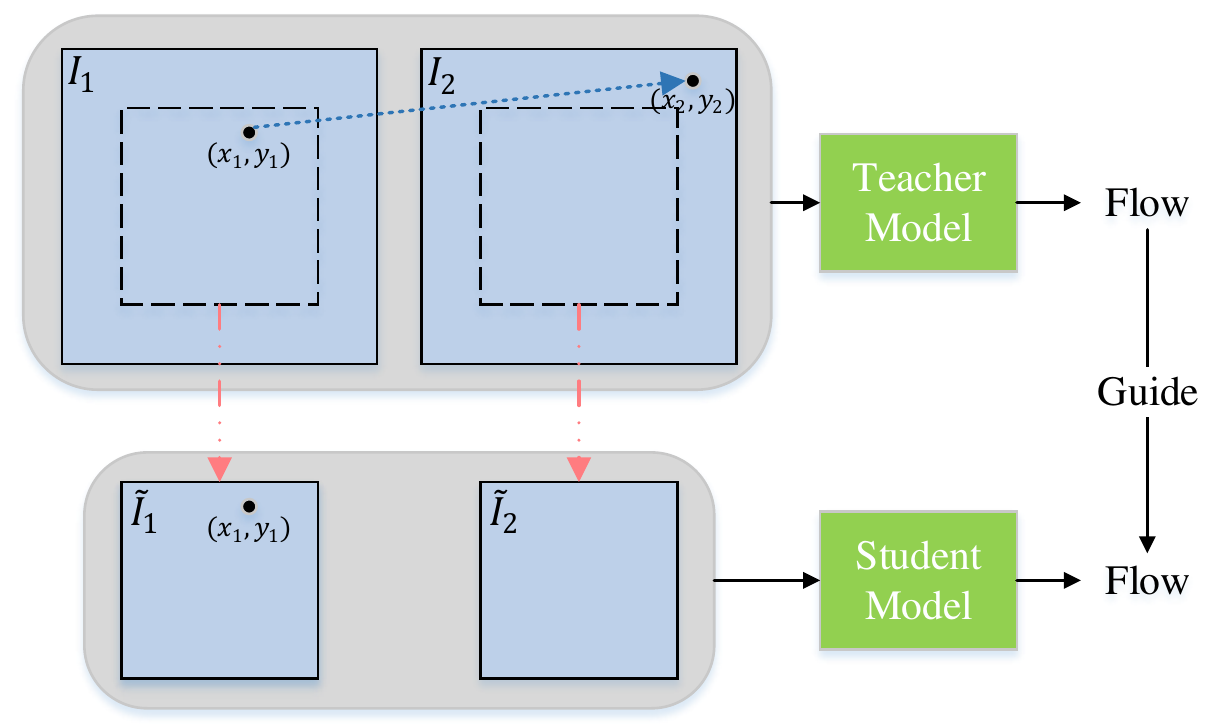}
 \caption{Data distillation illustration. We use the optical flow predictions from our teacher model to guide the learning of our student model.
}
 \label{ToyExample}
 \end{figure}

Is it possible to learn optical flow in a data-driven way, while not using any ground truth at all? In this paper, we address this issue by a data distilling approach. Our algorithm optimizes two models, a teacher model and a student model (as shown in Figure~\ref{ToyExample}). We train the teacher model  to estimate optical flow for  non-occluded pixels (e.g., $(x_1, y_1)$ in $I_1$). Then, we  hallucinate  flow  occlusion by cropping patches from original images  (pixel $(x_1, y_1)$ now becomes occluded in $\widetilde{I}_1$).
Predictions from our teacher model are  used as annotations to directly guide the student network to learn optical flow. Both networks share the identical architecture, and are trained end-to-end with simple loss functions. The student network is used to produce optical flow at test time, and runs at real time.

\begin{figure*}[th]
\centering
\includegraphics[width=0.9\textwidth]{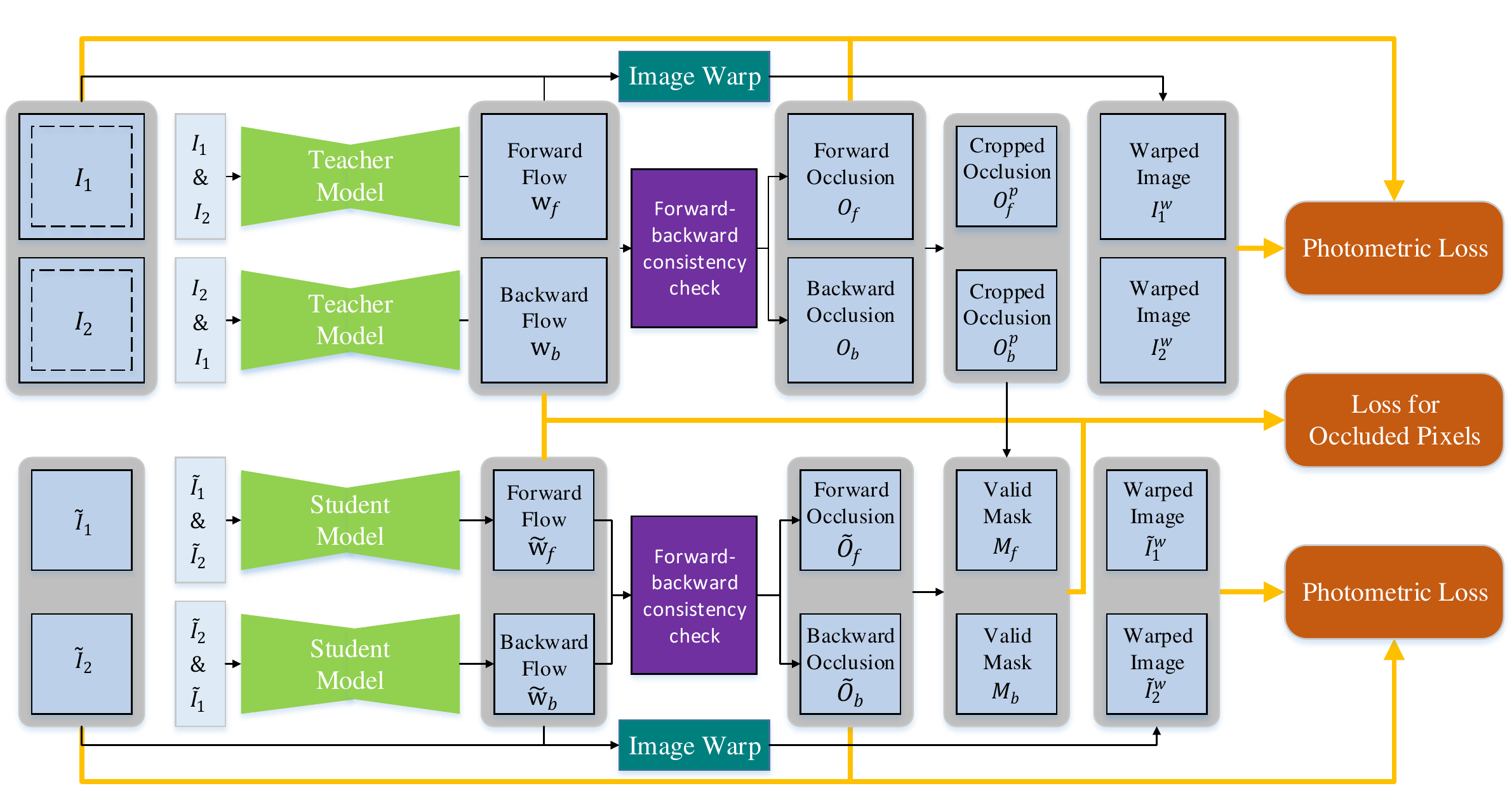}
\caption{Framework overview of DDFlow. Our teacher model and student model have identical network structures.
We train the teacher model  with  a photometric loss $L_p$ for non-occluded pixels.
The student model is trained with both $L_p$  and  $L_o$, a loss  for occluded pixels.
$L_o$ only functions on  pixels that are non-occluded in  original images but occluded in cropped patches (guided by \emph{Valid Mask $M_f$, $M_b$} ). During testing, only the student model is used.}
\label{ModelStrcuture}
\end{figure*}
The resulted self-training approach yields the highest accuracy among all unsupervised learning methods. At the time of writing, our method outperforms all published unsupervised flow methods on the Flying Chairs, MPI Sintel, KITTI 2012 and  2015 benchmarks. More notably, our method achieves a Fl-noc error of 4.57\% on KITTI 2012,  a Fl-all error of 14.29\% on KITTI 2015,  even outperforming several recent fully supervised methods which are fine-tuned for each dataset \cite{dosovitskiy2015flownet,ranjan2017optical,bailer2017cnn,zweig2017interponet,XRK2017}.

\begin{figure*}[th]
\centering
\includegraphics[width=0.92\textwidth]{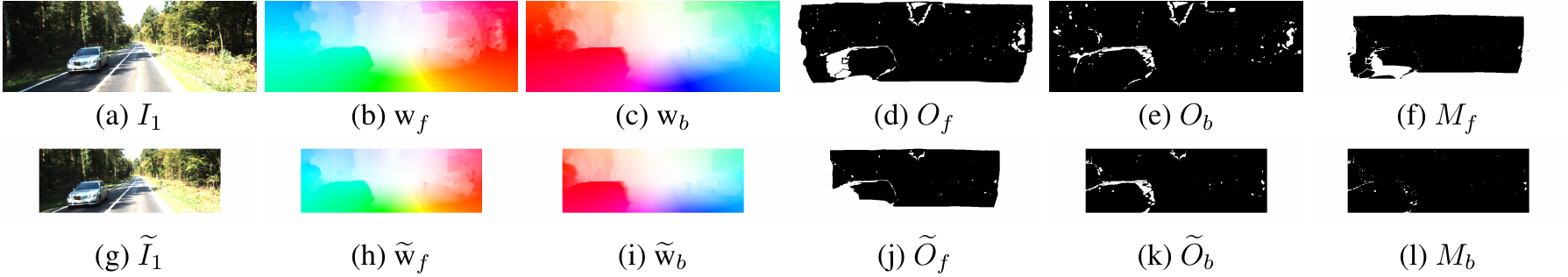}
\caption{Example intermediate results from DDFlow on KITTI. (a) is the first input image; (b,c) are forward and backward flow; (d,e) are forward and backward occlusion maps. (g) is the cropped patch of  (a); (h,i,j,k) are the corresponding forward flow, backward flow, forward occlusion map and backward occlusion map respectively. (f,l) are forward and backward valid mask, where  1 means the pixel is occluded in (g) but non-occluded in (a), 0 otherwise.}
\label{FlowWithPatchExample}
\end{figure*}

\section{Related Work}
Optical flow estimation has been a long-standing challenge in computer vision.
Early variational approaches \cite{horn1981determining,sun2010secrets} formulate it as an energy minimization problem based on brightness constancy and spatial smoothness. Such methods are effective for small motion, but  tend to fail when displacements are large.

Later, \cite{brox2011large,weinzaepfel2013deepflow} integrate feature matching  to tackle this issue. Specially, they find sparse feature correspondences to   initialize flow estimation and further refine it  in a pyramidal coarse-to-fine manner. The seminal work EpicFlow \cite{revaud2015epicflow} interpolates dense flow from sparse matches and has become a widely used post-processing pipeline. Recently, \cite{bailer2017cnn,XRK2017} use convolutional neural networks  to learn a  feature embedding for better matching  and have demonstrated superior performance. However,  all of these classical methods are often time-consuming, and their modules usually involve special tuning for different datasets.

The success of deep neural networks has motivated the development of  optical flow learning methods.
The pioneer work is FlowNet \cite{dosovitskiy2015flownet}, which takes two consecutive images as input and outputs a dense optical flow map. The following FlowNet 2.0 \cite{ilg2017flownet} significantly improves accuracy by stacking  several basic FlowNet modules together, and iteratively refining them. SpyNet \cite{ranjan2017optical} proposes to warp images at multiple scales to handle large displacements, and  introduces a compact spatial pyramid network to predict optical flow. Very recently, PWC-Net \cite{sun2018pwc} and LiteFlowNet \cite{hui18liteflownet} propose to warp features extracted from CNNs rather than warp images over different scales. They achieve state-of-the-art results while keeping a much smaller model size.
Though promising performance has been achieved, these methods require a large amount of labeled training data, which is particularly difficult to obtain for optical flow.

As a result, existing end-to-end deep learning based approaches \cite{dosovitskiy2015flownet,mayer2016large,Janai2018ECCV} turn to utilize synthetic  datasets  for pre-training.
Unfortunately, there usually exists a large domain gap between the distribution of synthetic datasets and natural scenes \cite{liu2008human}. Existing networks \cite{dosovitskiy2015flownet,ranjan2017optical} trained only on synthetic data turn to overfit, and  often perform poorly when directly evaluated on real sequences.

One promising direction is to develop unsupervised learning approaches.
\cite{jason2016back,ren2017unsupervised} construct loss functions based on brightness constancy and spatial smoothness. Specifically, the target image is warped according to the predicted flow, and then the difference between the reference image and the warped image is optimized using a photometric loss. Unfortunately, this loss  would provide misleading information when the pixels are occluded.

Very recently, \cite{Meister:2018:UUL,wang2018occlusion} propose to first reason occlusion map and then exclude those occluded pixels when computing the photometric difference. Most recently, \cite{Janai2018ECCV} introduce an unsupervised framework to estimate optical flow using a multi-frame formulation with temporal consistency. This method utilizes more data with more advanced occlusion reasoning, and hence  achieves more accurate results. However, all these unsupervised learning methods rely on hand-crafted energy terms to guide optical flow estimation,  lacking key  capability to learn optical flow of occluded pixels. As a consequence, the performance is still a large gap compared with state-of-the-art supervised methods.

To bridge this gap, we propose to perform
knowledge distillation from unlabeled data, inspired by \cite{hinton2015distilling,radosavovic2017data}
which performed knowledge distillation from multiple models or labeled data.
In contrast to previous knowledge distillation methods, we do not use any human annotations.
 Our idea is to generate annotations on unlabeled data using a
model trained with a classical
 optical flow energy, and then
retrain the model using those extra generated annotations.
This yields a simple yet effective method to learn optical flow for occluded pixels in a totally unsupervised manner.

\section{Method}

We first illustrate  our learning framework in Figure~\ref{ModelStrcuture}. We simultaneously train two CNNs (a teacher model and a student model) with the same structure. The teacher model is employed to predict  optical flow for non-occluded pixels and the student model is used to predict  optical flow of both non-occluded and occluded pixels. During testing time, only the student model is used to produce optical flow. Before describing our method, we define our notations as follows.

\subsection{Notation}
For our teacher model, we denote $I_1$, $I_2$ $\in\mathbb{R}^{H \times W \times 3}$ for two consecutive RGB images, where $H$ and $W$ are height and width respectively. Our goal is to estimate a forward optical flow $\text{w}_{f} \in \mathbb{R}^{H \times W \times 2} $ from $I_1$ to $I_2$. After obtaining  $\text{w}_{f}$, we can warp $I_2$ towards $I_1$ to get a warped image $I_2^w$. Here, we  also estimate a backward optical flow $\text{w}_b$ from $I_2$ to $I_1$ and a backward warp image $I_1^w$. Since there are many cases where one pixel is only visible in one image but not visible in the other image, namely occlusion, we denote $O_f$, $O_b$ $\in\mathbb{R}^{H \times W \times 1}$ as the forward and backward occlusion map respectively. For $O_f$ and $O_b$, value 1 means that the pixel in that location is occluded, while value 0 means not occluded.

 Our student model follows similar notations.
 We distill consistent predictions ($\text{w}_{f}$ and $O_f$) from our teacher model,  and crop patches on the original images to hallucinate occlusion.
 Let $\widetilde{I}_1$, $\widetilde{I}_2$, $\text{w}_f^p$, $\text{w}_b^p$, $O_f^p$ and $O_b^p$ denote the cropped image patches of $I_1$, $I_2$, $\text{w}_f$, $\text{w}_b$, $O_f$ and $O_b$ respectively. The cropping size is $h \times w$, where $h < H$, $w < W$.

The student network takes $\widetilde{I}_1$, $\widetilde{I}_2$ as input, and produces a forward and backward flow, a warped image, a occlusion map $\widetilde{\text{w}}_{f}$, $\widetilde{\text{w}}_{b}$, $\widetilde{I}_2^w$, $\widetilde{I}_1^w$, $\widetilde{O}_{f}$, $\widetilde{O}_{b}$ respectively.

 After obtaining $O_{f}^p$ and $\widetilde{O}_{f}$, we  compute another mask $M_f$, where value 1 means the pixel is occluded in image patch $\widetilde{I}_1$ but non-occluded in the original image $I_1$. The backward mask $M_b$ is computed in the same way. Figure~\ref{FlowWithPatchExample} shows  a real example for each notation used in DDFlow.

\begin{figure*}[th]
\centering
\includegraphics[width=0.92\textwidth]{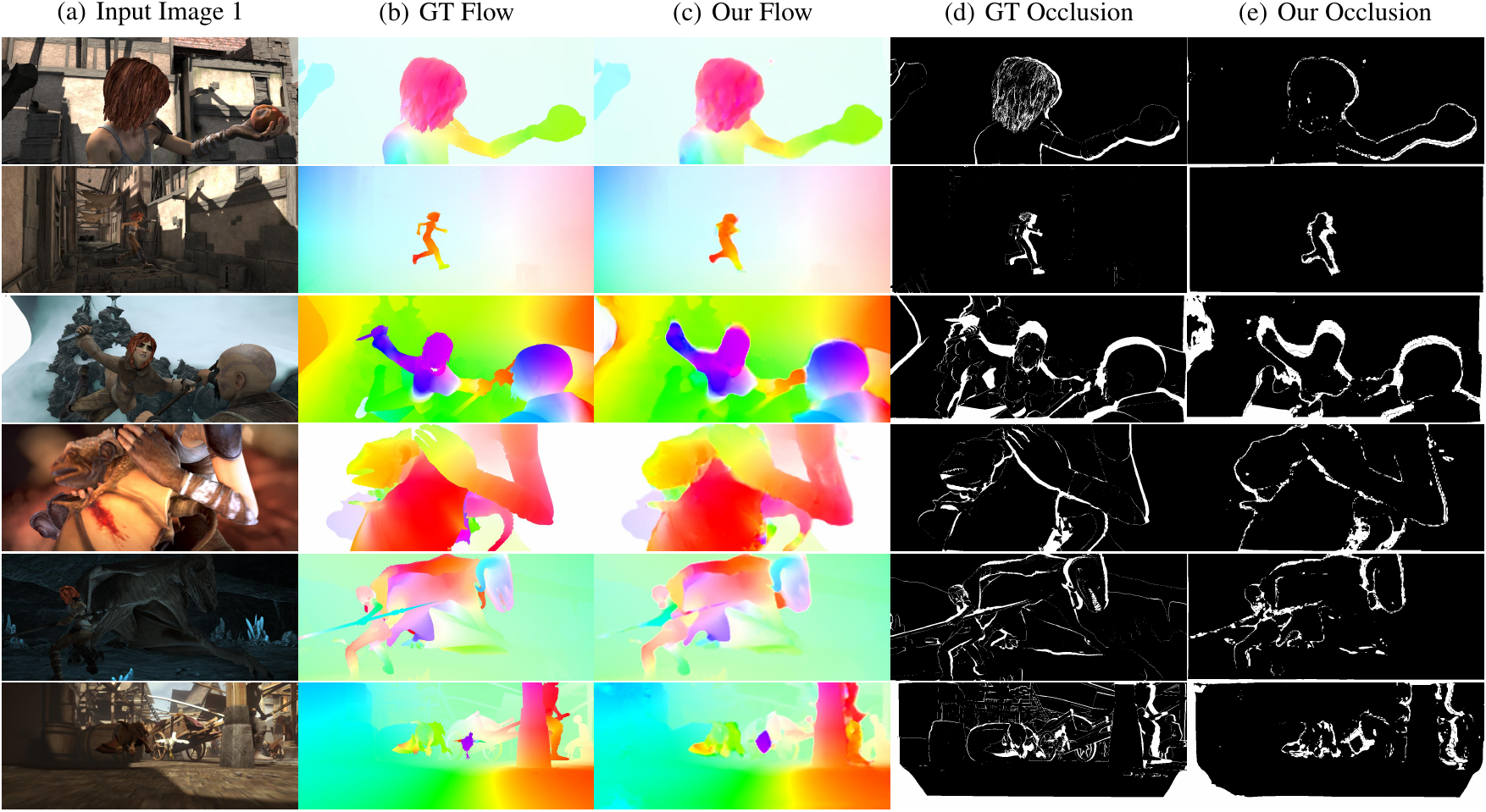}
\caption{Sample results on Sintel datasets. The first three rows are from Sintel Clean, while the last three are from Sintel Final. Our method  estimates accurate optical flow and reliable occlusion maps.}
\label{CompareSintel}
\end{figure*}

\begin{table*}[t]
\centering
\resizebox{0.96\textwidth}{!}{
\begin{tabular}{ c  l  c  c  c  c  c  c  c  c  c  c }
 \toprule
   &   \multirow{2}{*}{Method} & Chairs & \multicolumn{2}{c}{Sintel Clean} & \multicolumn{2}{c}{Sintel Final} & \multicolumn{3}{c}{KITTI 2012} & \multicolumn{2}{c}{KITTI 2015} \\
   \cmidrule(l{3mm}r{3mm}){4-5}    \cmidrule(l{3mm}r{3mm}){6-7} \cmidrule(l{3mm}r{3mm}){8-10}    \cmidrule(l{3mm}r{3mm}){11-12}
                   &  & test  & train & test & train & test & train & test & Fl-noc & train & Fl-all \\
  \midrule
   \multirow{8}{*}{\rotatebox[origin=c]{90}{Supervise}}
   & FlowNetS~\cite{dosovitskiy2015flownet}    & 2.71      & 4.50      & 7.42      & 5.45      & 8.43      & 8.26      & --    & -- & --     & --      \\
   & FlowNetS+ft~\cite{dosovitskiy2015flownet} & --        & (3.66)    & 6.96      & (4.44)    & 7.76      & 7.52      & 9.1   & -- & --     & --      \\
   & SpyNet~\cite{ranjan2017optical}           & 2.63      & 4.12      & 6.69      & 5.57      & 8.43      & 9.12      & --    & -- & --     & --      \\
   & SpyNet+ft~\cite{ranjan2017optical}        & --        & (3.17)    & 6.64      & (4.32)    & 8.36      & 8.25      & 10.1  & 12.31\% & --     & 35.07\%  \\
   & FlowNet2~\cite{ilg2017flownet}            & --        & 2.02      & 3.96      & 3.14      & 6.02      & 4.09      & --    & -- & 10.06  & --      \\
   & FlowNet2+ft~\cite{ilg2017flownet}         & --        & (1.45)    & 4.16      & (2.01)    & 5.74      & (1.28)    & 1.8   & 4.82\% & (2.3)  & 11.48\% \\
   & PWC-Net~\cite{sun2018pwc}                 & \textbf{2.00}& 3.33   & --        & 4.59      & --        & 4.57      & --    & -- & 13.20  & --      \\
   & PWC-Net+ft~\cite{sun2018pwc}              & --&\textbf{(1.70)} &\textbf{3.86}& \textbf{(2.21)}&\textbf{5.13}&\textbf{(1.45)}&\textbf{1.7}&\textbf{4.22\%} &\textbf{(2.16)} & \textbf{9.60\%}  \\
  \midrule
   \multirow{11}{*}{\rotatebox[origin=c]{90}{Unsupervise}}
   & BackToBasic+ft~\cite{jason2016back}       & 5.3       & --        & --        & --        & --        & 11.3      & 9.9   & -- & --     & --      \\
   & DSTFlow+ft~\cite{ren2017unsupervised}     & 5.11      & (6.16)    & 10.41     & (6.81)    & 11.27     & 10.43     & 12.4  & -- & 16.79  & 39\%    \\
   & UnFlow-CSS+ft~\cite{Meister:2018:UUL}     & --        & --        & --        & (7.91)    & 10.22     & 3.29      & --    & -- & 8.10   & 23.30\% \\
   & OccAwareFlow~\cite{wang2018occlusion}     & 3.30      & 5.23      & 8.02      & 6.34      & 9.08      & 12.95     & --    & -- & 21.30  & --      \\
   & OccAwareFlow+ft-Sintel~\cite{wang2018occlusion} &3.76 & (4.03)    & 7.95      & (5.95)    & 9.15      & 12.9      & --    & -- & 22.6   & --      \\
   & OccAwareFlow-KITTI~\cite{wang2018occlusion} & --      & 7.41      & --        & 7.92      & --        & 3.55      & 4.2   & -- & 8.88   & 31.2\%  \\
   & MultiFrameOccFlow-Hard+ft~\cite{Janai2018ECCV}& --    & (6.05)    & --        & (7.09)    & --        & --        & --    & -- & 6.65   & --      \\
   & MultiFrameOccFlow-Soft+ft~\cite{Janai2018ECCV}& --    & (3.89)    & 7.23      & (5.52)    & 8.81      & --        & --    & -- & 6.59   & 22.94\% \\
   \cline{2-12}
   & DDFlow                                      &\textbf{2.97}&  3.83     & --      & 4.85     & --       & 8.27        & --    & -- & 17.26  &  --     \\
   & DDFlow+ft-Sintel                            & 3.46      &\textbf{(2.92)}&\textbf{6.18}&\textbf{(3.98)}&\textbf{7.40}& 5.14  & -- & -- & 12.69  &  --     \\
   & DDFlow+ft-KITTI                             & 6.35      &  6.20     & --        & 7.08  & -- &\textbf{2.35} & \textbf{3.0} & \textbf{4.57\%} &\textbf{5.72}&\textbf{14.29\%} \\
 \bottomrule \end{tabular} }
\caption{Comparison to state-of-the-art optical flow estimation methods. All numbers are   EPE except for the last column of  KITTI 2012 and KITTI 2015 test sets, where we report percentage of erroneous pixels (Fl). Missing entries (-) indicate that the results are not reported for the respective method. Parentheses mean that the training is performed on the same dataset. Bold fonts highlight the best results among supervised and unsupervised methods respectively. Note that MultiFrameOccFlow \cite{Janai2018ECCV} utilizes multiple frames, while all other methods use only two consecutive frames.}
\label{QuantitativeResult}
\end{table*}

\subsection{Network Architecture}
In principle, DDFlow can use any backbone network to learn optical flow.
We select PWC-Net \cite{sun2018pwc} as our backbone network due to its remarkable performance and compact model size. PWC-Net learns 7-level feature representations for two input images, and gradually conducts feature warping and cost volume construction  from  the last level  to the third level. As a result, the output resolution of flow map is a quarter of the original image size. We upsample the output flow to the full resolution using bilinear interpolation.
To train  two networks simultaneously in a totally unsupervised way, we normalize features when constructing cost volume, and  swap the image pairs in our input to produce both forward and backward flow.

We use the identical network architecture for our teacher and student model. The only difference between them is to train each with different input data and loss functions. Next, we discuss how to generate such data, and construct loss functions for each model in detail.

\subsection{Unlabeled Data Distillation}
For prior unsupervised optical flow learning methods, the only guidance is a photometric loss which measures the difference between the reference image and the warped target image. However, photometric loss makes no sense for  occluded pixels.
To tackle this issue, We distill predictions from our teacher model, and use them to generate input/output data for our student model. Figure~\ref{ToyExample} shows a toy example for our data distillation idea.

 Suppose  pixel ($x_2$, $y_2$) in $I_2$ is the corresponding pixel of ($x_1$, $y_1$) in $I_1$.
 Given ($x_1$, $y_1$) is non-occluded, we can use the classical photometric loss to find its optical flow using our teacher model. Now, if we crop image patches $\widetilde{I}_1$ and $\widetilde{I}_2$,  pixel ($x_1$, $y_1$) in $\widetilde{I}_1$ becomes occluded, since there is no corresponding pixel in $\widetilde{I}_2$ any more.
 Fortunately, the optical flow prediction for ($x_1$, $y_1$) from our teacher model is still there. We then directly use this prediction as annotation to guide the student model to learn optical flow for the occluded pixel ($x_1$, $y_1$) in $\widetilde{I}_1$.
 This is  the key intuition behind DDFlow.

Figure~\ref{ModelStrcuture} shows the main data flow for our approach.
To make full use of the input data, we compute both forward and backward flow $\text{w}_{f}$, $\text{w}_{b}$ for the original frames, as well as their warped images $I_1^w$, $I_2^w$.
We also estimate two occlusion maps $O_f$, $O_w$ by checking forward-backward consistency.
The  teacher model is trained with a photometric loss, which minimizes a warping error using $I_1$, $I_2$,
$O_f$, $O_w$, $I_1^w$, $I_2^w$.
This  model produces accurate optical flow predictions for non-occluded pixels in $I_1$ and $I_2$.

For our student model, we randomly crop image patches $\widetilde{I}_1$, $\widetilde{I}_2$ from $I_1$, $I_2$, and we compute  forward and backward flow $\widetilde{\text{w}}_{f}$, $\widetilde{\text{w}}_{b}$ for them.
A similar photometric loss is employed for the non-occluded pixels in $\widetilde{I}_1$ and $\widetilde{I}_2$.
In addition, predictions from our teacher model are employed as output  annotations to guide those pixels occluded in cropped image patches but non-occluded in original images. Next, we discuss how to construct all the loss functions.

\subsection{Loss Functions}
Our loss functions include two components: photometric loss $L_p$ and  loss for occluded pixels $L_o$. Optionally, smoothness losses can also be added. Here, we focus on the above two loss terms for simplicity. For the teacher model, only $L_p$ is used to estimate the flow of non-occluded pixels, while for student model, $L_p$ and $L_o$ are both employed to estimate the optical flow of non-occluded and occluded pixels.

\textbf{Occlusion Estimation.} Our occlusion detection is based on the forward-backward consistency prior \cite{sundaram2010dense,Meister:2018:UUL}. That is, for non-occluded pixels, the forward flow should be the inverse of the backward flow at the corresponding pixel in the second image. We consider pixels as occluded when the mismatch between forward flow and backward flow is too large or the flow is out of image boundary $\Omega$. Take a forward occlusion map as an example, we  first compute the reversed forward flow $\hat{\text{w}}_f = \text{w}_b({\textbf{p}+\text{w}_f}(\textbf{p}))$, where $\textbf{p} \in \Omega$. A pixel is considered occluded if either of the following constraints is violated:
\begin{equation}
\left\{
             \begin{array}{lr}
             |\text{w}_f + \hat{\text{w}}_f|^2 < \alpha_1 (|\text{w}_f|^2+|\hat{\text{w}}_f|^2) + \alpha_2, &  \\
             \textbf{p}+\text{w}_f(\textbf{p}) \notin \Omega, &
             \end{array}
\right.
\end{equation}
where we set $\alpha_1$ = 0.01, $\alpha_2$ = 0.05 for all our experiments. Backward occlusion maps are computed in the same way.

\textbf{Photometric Loss.} The photometric loss is based on the brightness constancy assumption, which measures the difference between the reference image and the warped target image. It is only effective for   non-occluded pixels. We define a simple loss as follows:
\begin{multline}
	L_p = \sum{\psi(I_1-I_2^{w}) \odot (1 - O_f)} / \sum{(1 - O_f)} \\
	+ \sum{\psi(I_2-I_1^{w}) \odot (1-O_b)} / \sum(1 - O_b)
  \label{eq:nocloss}
\end{multline}
where $\psi(x) = (|x|+\epsilon)^q$ is a robust loss function, $\odot$ denotes the element-wise multiplication. During our experiments, we set $\epsilon=0.01$, $q=0.4$. Our teacher model  only minimizes this loss.

\textbf{Loss for Occluded Pixels.} The key element in unsupervised learning is the loss for occluded pixels.
In contrast to existing loss functions relying on smoothing prior to constrain  flow estimation, our loss is purely data-driven.
This enables us to directly learn from real data, and produce more accurate flow.
To this end, we define our loss on  pixels that are occluded in the cropped patch but non-occluded in the original image.
Then, supervision is generated using  predictions  of the original image from our teacher model, which produces reliable optical flow for non-occluded pixels.

To find these pixels, we  first compute a valid mask $M$ representing the pixels that are occluded in the cropped image but non-occluded in the original image:
\begin{equation}
M_f = \text{clip}(\widetilde{O}_f-O_f^{p}, 0, 1)
\end{equation}
Backward mask $M_b$  is computed in the same way. Then we define our   loss for occluded pixels in the following,
\begin{multline}
	L_o = \sum{\psi(\text{w}_f^{p}-\widetilde{\text{w}}_f) \odot M_f} / \sum{M_f} \\
	+ \sum{\psi(\text{w}_b^{p}-\widetilde{\text{w}}_b) \odot M_b} / \sum{M_b}
  \label{eq:occloss}
\end{multline}
We use the same robust loss function $\psi(x)$ with the same parameters defined in Eq.~\ref{eq:nocloss}. Our student model minimizes the simple combination $L_p$+$L_o$.

\section{Experiments}

\begin{figure*}[th]
\centering
\includegraphics[width=0.92\textwidth]{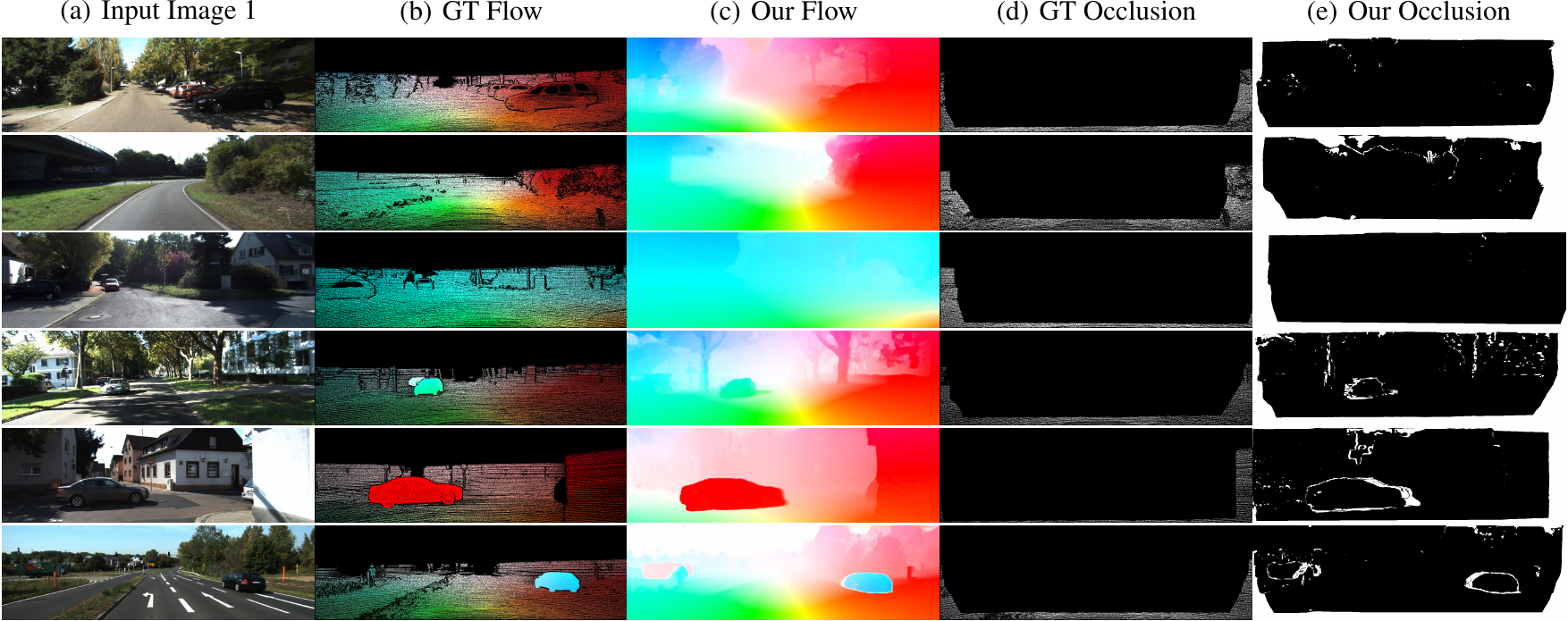}
\caption{Example results on KITTI datasets. The first three rows are from KITTI 2012, and the last three are from KITTI 2015. Our method estimates accurate optical flow and reliable occlusion maps. Note that on KITTI datasets, the occlusion masks are sparse and only contain pixels moving out of the image boundary.}
\label{CompareKITTI}
\end{figure*}

We evaluate DDFlow on standard optical flow benchmarks including Flying Chairs \cite{dosovitskiy2015flownet}, MPI Sintel \cite{butler2012naturalistic},  KITTI 2012\cite{geiger2012we}, and KITTI 2015 \cite{menze2015object}. We compare our results with  state-of-the-art unsupervised methods including BackToBasic\cite{jason2016back}, DSTFlow\cite{ren2017unsupervised}, UnFlow\cite{Meister:2018:UUL}, OccAwareFlow\cite{wang2018occlusion} and MultiFrameOccFlow\cite{Janai2018ECCV}, as well as fully supervised learning methods including FlowNet\cite{dosovitskiy2015flownet}, SpyNet\cite{ranjan2017optical}, FlowNet2\cite{ilg2017flownet} and PWC-Net\cite{sun2018pwc}. Note that MultiFrameOccFlow \cite{Janai2018ECCV} utilizes multiple frames as input, while all other methods use only two consecutive frames. 
To ensure reproducibility and advance further innovations, we  make our code and models publicly available our our project website.

\subsection{Implementation Details}

\textbf{Data Preprocessing.}
We preprocess the image pairs using census transform \cite{zabih1994non}, which is proved to be robust for optical flow estimation \cite{hafner2013census}. We find that this simple procedure can indeed improve the performance of unsupervised optical flow estimation, which is consistent with \cite{Meister:2018:UUL}.

\textbf{Training procedure.}
For all our experiments, we use the same network architecture and train our model using Adam optimizer \cite{kingma2014adam} with $\beta_1$ =0.9 and $\beta_2$=0.999. For all datasets, we set batch size as 4. For all individual experiments, we use a initial learning rate of  1e-4, and it decays half every 50k iterations. For data augmentation, we only use random cropping, random flipping, and random channel swapping. Thanks to the simplicity of our loss functions, there is no need to tune hyper-parameters.

Following prior work, we first pre-train DDFlow on Flying Chairs. We initialize our teacher network from random, and warm it up with   200k iterations using our  photometric loss without considering occlusion. Then, we add our occlusion detection check, and train the network with the photometric loss $L_p$ for another 300k iterations. After that, we initialize the student model with the weights from our teacher model, and train both the teacher model (with $L_p$) and the student model (with $L_p+L_o$) together for 300k iterations. This concludes our pre-training, and the student model is used for future fine-tuning.

We use the same fine-tuning procedure for  all Sintel and  KITTI datasets. First, we initialize the teacher network using the pre-trained student model from Flying Chairs, and train it
for 300k iterations. Then, similar to pre-training on Flying Chairs, the student network is initialized with the new teacher model, and both networks are  trained together for another 300k iterations.
The student model  is used during our evaluation.

\textbf{Evaluation Metrics.}
We consider two widely-used metrics to evaluate optical flow estimation and one metric of occlusion evaluation:
average endpoint error (EPE), percentage of erroneous pixels (Fl),  harmonic average of the precision and recall (F-measure).
We also report the results of EPE over non-occluded pixels (NOC)  and occluded pixels (OCC) respectively.
EPE is the ranking metric on MPI Sintel benchmark, and Fl is the ranking metric on KITTI benchmarks.

\begin{table*}[t]
\centering
\resizebox{0.96\textwidth}{!}{
\begin{tabular}{ c c c  c  c c c  c c c  c c c  c c c }
 \toprule
	Occlusion   & Census & Data   &  Chairs  & \multicolumn{3}{c}{Sintel Clean}& \multicolumn{3}{c}{Sintel Final} & \multicolumn{3}{c}{KITTI 2012} & \multicolumn{3}{c}{KITTI 2015} \\
  \cmidrule(l{3mm}r{3mm}){5-7}   \cmidrule(l{3mm}r{3mm}){8-10}   \cmidrule(l{3mm}r{3mm}){11-13}   \cmidrule(l{3mm}r{3mm}){14-16}
	Handling  	& Transform  & Distillation &   ALL   &  ALL   & NOC   &  OCC    &   ALL  &   NOC  &  OCC   &   ALL &   NOC  &  OCC  &   ALL  &   NOC   &   OCC \\
  \midrule
	\xmark     	& \xmark    & \xmark     &   4.06   & (5.05) & (2.45)&  (38.09)& (7.54) & (4.81) &(42.46) &  10.76&  3.35  & 59.86 & 16.85  &   6.45  &  82.64\\
	\cmark 		& \xmark 	& \xmark     &   3.95   & (4.45) & (2.16)&  (33.48)& (6.56) & (4.12) &(37.83) &  6.67 &  1.94  & 38.01 & 12.42  &   5.67  &  60.59\\
	\xmark     	& \cmark    & \xmark     &   3.75   & (3.90) & (1.60)&  (33.31)& (5.23) & (2.80) &(36.35) &  8.66 &  1.47  & 56.24 & 14.04  &   4.06  &  77.16\\
	\cmark 		& \cmark 	& \xmark 	 &   3.24   & (3.37) & (1.34)& (29.36) & (4.47) & (2.32) &(31.86) & 4.50  &  1.10  & 27.04 &  8.01  &  3.02   &  42.66\\
	\cmark	  	& \cmark    & \cmark 	 &   \textbf{2.97}   & \textbf{(2.92)} & \textbf{(1.27)}& \textbf{(23.92)} & \textbf{(3.98)} & \textbf{(2.21)} &
	\textbf{(26.74)}& \textbf{2.35}  &  \textbf{1.02}  & \textbf{11.31} &  \textbf{5.72}  &  \textbf{2.73}   &   \textbf{24.68}\\
 \bottomrule
\end{tabular} } 
\caption{Ablation study. We compare the results of EPE over all pixels (ALL), non-occluded pixels (NOC) and occluded pixels (OCC) under different settings. Bold fonts highlight the best results.}
\label{Ablation}
\end{table*}

\begin{table}[th]   \centering
\resizebox{0.45\textwidth}{!}{
\begin{tabular}{ l c c c c }
 \toprule 	\multirow{2}{*}{Method}                      & Sintel  & Sintel  & KITTI    & KITTI \\
			                      & Clean   & Final   & 2012     & 2015  \\
  \midrule
 	MODOF                         &   --    & 0.48    &  --      & --   \\
 	OccAwareFlow-ft               &  (0.54) & (0.48)  & \textbf{0.95$^*$} & 0.88$^*$   \\
 	MultiFrameOccFlow-Soft+ft     &  (0.49) & (0.44)  &  --      & \textbf{0.91$^*$} \\
 	Ours                          &  \textbf{(0.59)} & \textbf{(0.52)}  & 0.94$^*$ & 0.86$^*$ \\
  \bottomrule \end{tabular}}
\caption{Comparison to state-of-the-art occlusion estimation methods.  $^*$ marks cases where  the occlusion map is sparse and only the annotated pixels are considered.
} \label{OcclusionEstimation}
\end{table}

\subsection{Comparison to State-of-the-art}
We compare our results with state-of-the art methods in Table \ref{QuantitativeResult}. As we can see, our approach, DDFlow, outperforms all existing unsupervised flow learning methods on all datasets.
 On the test set of Flying Chairs, our EPE is better than all prior results, decreasing from previous state-of-the-art 3.30 to 2.97. More importantly, simply evaluating our model only pre-trained on Flying Chairs, DDFlow  achieves EPE=3.83 on Sintel Clean and EPE=4.85 on Sintel Final, which are even better than the results from state-of-the-art unsupervised methods \cite{wang2018occlusion,Janai2018ECCV} finetuned specifically for Sintel. This is remarkable,  as it shows the great generalization capability of DDFlow.

After we finetuned DDFlow using frames from the Sintel training set, we achieved an EPE=7.40 on the Sintel Final testing benchmark, improving the best  prior result (EPE=8.81 from \cite{Janai2018ECCV}) by a relative margin of 14.0~\%.
Similar improvement (from 7.23 to 6.18) is also observed on the Sintel Clean testing benchmark.
 Our model is even better than some supervised methods including \cite{dosovitskiy2015flownet} and \cite{ranjan2017optical}, which are finetuned on Sintel using ground truth annotations. Figure~\ref{CompareSintel} shows sample DDFlow results from  Sintel,  comparing our optical flow estimations and occlusion masks with the ground truth.

 On KITTI dataset, the improvement from DDFlow is even more significant. On KITTI 2012 testing set, DDFlow yields an  EPE=3.0, 28.6~\% lower than the best existing counterpart (EPE=4.2 from \cite{wang2018occlusion}). For the ranking measurement on KITTI 2012, we achieve Fl-noc=4.57~\%, even better than the result (4.82~\%) from the well-known FlowNet 2.0.
For KITTI 2015, DDFlow performs particularly well. The Fl-all from DDFlow reaches 14.29\%, not only
better than the best unsupervised method by a large margin (37.7~\% relative improvement), but also outperforming
several recent fully supervised learning methods including \cite{ranjan2017optical,bailer2017cnn,zweig2017interponet,XRK2017}.
Example results from  KITTI 2012 and 2015 can be seen in  Figure~\ref{CompareKITTI}.

\subsection{Occlusion Estimation}
Next, we evaluate our occlusion estimation on both Sintel and KITTI dataset. We compare our method with MODOF\cite{xu2012motion}, OccAwareFlow-ft\cite{wang2018occlusion}, MultiFrameOccFlow-Soft+ft\cite{Janai2018ECCV} using F-measure. Note KITTI datasets only have sparse occlusion map.

As shown in  Table~\ref{OcclusionEstimation}, our method achieve best occlusion estimation performance on Sintel Clean and Sintel Final datasets over all competing methods. On KITTI dataset, the ground truth occlusion masks only contain pixels moving out of the image boundary. However, our method will also estimate the occlusions within the image range. Under such settings, our method can achieve comparable performance.

\subsection{Ablation Study}
We conduct a thorough ablation analysis for different components of DDflow. We report our findings in Table \ref{Ablation}.

\textbf{Occlusion Handling.} Comparing the first row and the second row, the third row and the fourth row, we can see that occlusion handling can improve the optical flow estimation performance over all pixels, non-occluded pixels and occluded pixels on all datasets. It is because that brightness constancy assumption does not hold for occluded pixels.

\textbf{Census Transform.} Census transform can compensate for illumination changes, which is robust for optical flow estimation and has been widely used in traditional methods. Comparing the first row and the third row, the second row and the fourth row, we can see that it indeed constantly improves the performance on all datasets.

\textbf{Data Distillation.} Since brightness constancy assumption does not hold for occluded pixels and there is no ground truth flow for occluded pixels, we introduce a data distillation loss to address this problem. As shown in the fourth row and the fifth row, occluded prediction can improve the performance on all datasets, especially for occluded pixels. EPE-OCC decreases from 29.36 to 23.93 (by 18.5~\%) on Sintel Clean, from 31.86 to 26.74 (by 16.1~\%) on Sintel Final dataset, from 27.04 to 11.31 (by 58.2~\%) on KITTI 2012  and from 42.66 to 24.68 (by 42.1~\%) on KITTI 2015. Such a big improvement demonstrates the effectiveness of DDFlow.

Our distillation strategy works particularly well near image boundary, since our teacher model can distill reliable labels for these pixels. For occluded pixels elsewhere, our method is not as effective, but still produces reasonable results to some extent. This is because we crop at random location for student model, which covers a large amount of occlusions. Exploring new ideas to cope with occluded pixels at any location can be a promising research direction in the future.

\section{Conclusion}
We have presented  a  data distillation approach to learn optical flow from unlabeled data.
We have shown that  CNNs can be self-trained to estimate optical flow, even for occluded pixels, without using any human annotations. To this end, we construct two networks. The predictions  from the teacher network are used as annotations to guide the student network to learn optical flow.  Our method, DDFlow, has achieved the highest accuracy among all prior unsupervised methods on all challenging optical flow benchmarks. 
Our work makes a step  towards distilling optical flow knowledge from unlabeled data.
Going forward, our  results suggest that our data distillation technique may be a promising direction for advancing other vision tasks like stereo matching \cite{zbontar2016stereo} or depth estimation \cite{eigen2014depth}.

\bibliographystyle{aaai}
\bibliography{DDFlow}

\end{document}